\pdfoutput=1

\documentclass[11pt]{article}

\usepackage[final]{acl}

\usepackage{times}
\usepackage{latexsym}

\usepackage[T1]{fontenc}

\usepackage[utf8]{inputenc}

\usepackage{microtype}

\usepackage{inconsolata}

\usepackage{graphicx}

\usepackage{booktabs}

\usepackage{enumitem, tcolorbox}
\AtBeginEnvironment{tcolorbox}{\small}
\tcbset{
    sharp corners,
    colback = white,
    before skip = 0.2cm,
    after skip = 0.5cm
}                           
\newtcolorbox{boxA}{
    boxrule = 1pt,
    colframe = black 
}

\title{Generating Media Background Checks for Automated Source Critical Reasoning}

\author{Michael Schlichtkrull \\
  School of Electronic Engineering and Computer Science \\
  Queen Mary University of London \\
  \texttt{m.schlichtkrull@qmul.ac.uk}\\}

\begin{document}
\maketitle
\begin{abstract}
Not everything on the internet is true. This unfortunate fact requires both humans and models to perform complex reasoning about credibility when working with retrieved information. In NLP, this problem has seen little attention. Indeed, retrieval-augmented models are not typically expected to distrust retrieved documents. Human experts overcome the challenge by gathering signals about the context, reliability, and tendency of source documents -- that is, they perform \textit{source criticism}. We propose a novel NLP task focused on finding and summarising such signals. We introduce a new dataset of 6,709 ``media background checks'' derived from Media Bias / Fact Check, a volunteer-run website documenting media bias. We test open-source and closed-source LLM baselines with and without retrieval on this dataset, finding that retrieval greatly improves performance. We furthermore carry out human evaluation, demonstrating that 1) media background checks are helpful for humans, and 2) media background checks are helpful for retrieval-augmented models.
\end{abstract}

\section{Introduction}

When humans perform knowledge-intensive reasoning, we are rarely able to rely on a single, authoritative source. Instead, we forage for multiple sources, evaluate their trustworthiness, and synthesize answers~\citep{potter2013media}. The basic task is to choose \textit{reliable} sources, to read them \textit{reliably}, and to combine them into \textit{reliable} narratives~\citep{howell2001reliable}. Best practice for epistemic experts, such as journalists and historians, is to rely on multiple sources, to present evidence of source tendency and reliability, and to explain source disagreements to readers~\citep{steensen_journalisms_2019}.  Search engines, acting as surrogate experts~\citep{simpson2013evaluating}, similarly enrich their results with knowledge-contexts that help users reason about tendency and trust~\citep{smith2019knowledge}.

 \begin{figure}[t]
\centering
\includegraphics[scale=0.8]{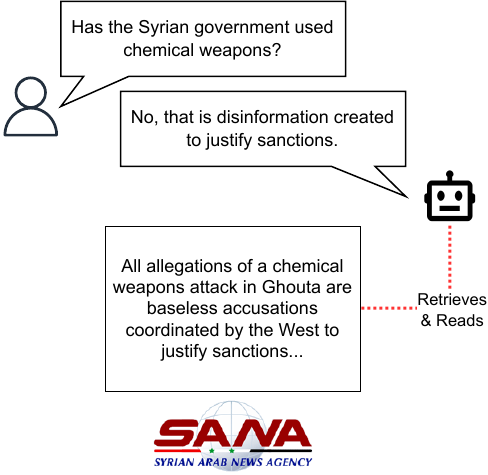}
\caption{
Retrieval-augmented NLP models can inadvertently misinform users if uncritically relying on retrieved documents from such untrustworthy sources. In preliminary experiments we found evidence of this occurring in practise: One popular search-augmented chatbot engaged in war-crimes denial after relying on Syrian state news to answer questions.
}
\label{fig:example}
\end{figure}

Source-critical reasoning has not received much attention in NLP. Even for settings with clearly disputed truth claims, such as fact-checking, studies typically assume a single, trustworthy source, e.g. Wikipedia~\citep{thorne-etal-2018-fever}, scientific journals~\citep{wadden-etal-2020-fact}, or search results~\citep{schlichtkrull2023averitec}. This is the case even when researchers propose to fully automate away human epistemic experts~\citep{schlichtkrull-etal-2023-intended}. Problems of trust, uncertainty, and disagreeing evidence are often mentioned in sections like ``broader impact''~\citep{lewis2020rag} or ``limitations''~\citep{schlichtkrull2023averitec}.

\begin{figure*}[ht]
\centering
\includegraphics[scale=0.75]{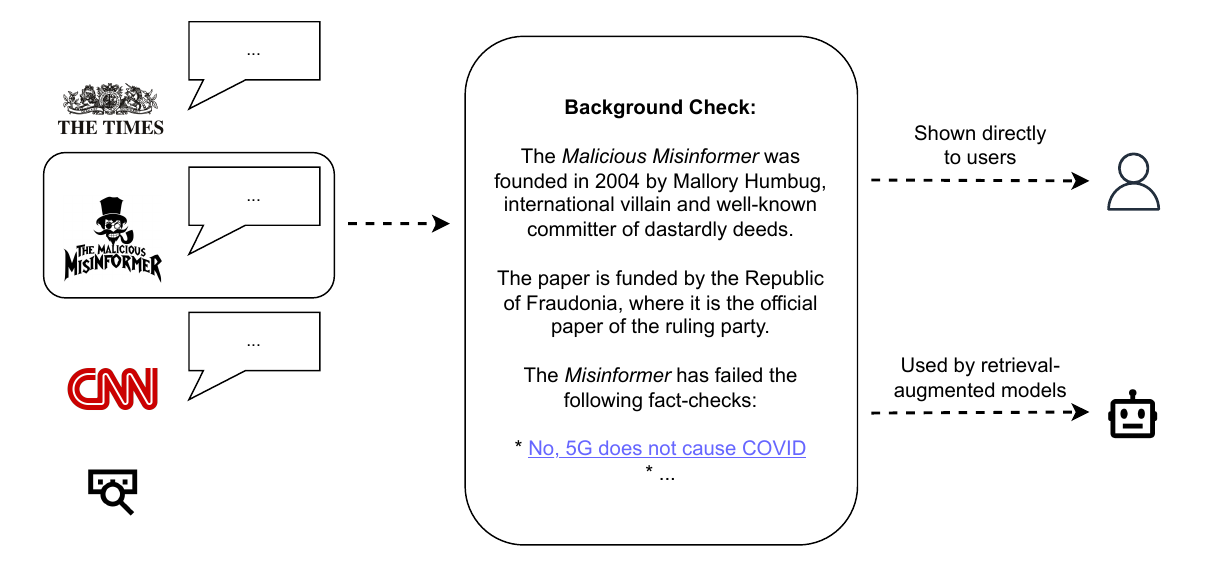}
\caption{\label{figure:background_check} We propose to generate \textit{Media Background Checks} (MBCs) that summarise indicators of trustworthiness and tendency. MBCs can be used, either by humans or by retrieval-augmented models, to determine which documents can be relied on for further reasoning, and to craft reliable narratives based on untrustworthy evidence.
}
\end{figure*}

Recently, large language models (LLMs) have been envisioned as alternatives to search~\citep{metzler2021rethinking}. This is risky: in assessments by domain experts, LLMs have been found to give definitive answers on subjects without adequate consensus
~\citep{peskoff-stewart-2023-credible}. 
One proposed solution is retrieval-augmentation, where models reason based on retrieved documents (e.g., from search) rather than weights~\citep{lewis2020rag}. However, it still falls on the user to analyse if the sources used can be trusted -- and LLMs often provide less knowledge-context than traditional search engines~\citep{shah2022situtating}. In the absence of other evidence, users often fall back on search rankings to determine trustworthiness; this ``Google-ization'' is an existing concern in journalism, where overreliance on search can cause a ``distortion of reality''~\citep{machill2009importance, overreliance}. 
LLMs can similarly misinform users if they rely uncritically on retrieved documents.  In preliminary experiments, we observed one popular model deny war crimes as a result of an untrustworthy retrieved document (see Figure~\ref{fig:example}).

When reasoning based on secondhand knowledge, even the most expert human is prisoner to their sources. Nevertheless, with awareness of these limitations, readers can rely on signals about the motivation and tendencies of their sources to create understanding; i.e., they can carry out source criticism~\citep{howell2001reliable, Godler2015, steensen_journalisms_2019}. Human knowledge experts are expected to assist their readers in this process. For example, member organisations of the International Fact-checking Consortium are required to \textit{``identify and disclose the \textbf{relevant interests} of the sources it uses where the reader might reasonably conclude those interests could influence the accuracy of the evidence provided''}~\citep{IFCN2016}. Similar to \citet{imperial2024standardize}, we argue that models should be aligned to best practises for human experts. Disclosing not just the names of sources, but also signals about their motivations and tendencies, is crucial as these indicators can be difficult to find for non-expert humans who may not be intimately familiar with each particular source.

In this paper, we provide building blocks for analysing the biases and tendencies of retrieved documents, taking the first steps towards source-critical NLP. We introduce a new task, \textbf{generating media background checks}. Media background checks (MBCs) are short statements that give context to sources, enabling critical analysis (see Figure~\ref{figure:background_check}). They cover topics like ownership, funding model, remit, known biases, and factuality signals such as previous failed fact-checks -- the same signals used by human experts in their analysis~\citep{howell2001reliable}. The ability to construct an MBC, i.e. to recall the necessary information, is a prerequisite for further source-critical reasoning. As we show in Section~\ref{section:human_evaluation}, MBCs can help both humans and models reason source-critically.

We create a dataset of 6,709 MBCs collected from the website Media Bias / Fact Check\footnote{https://mediabiasfactcheck.com/}, which publishes detailed reports on media organisations as part of their effort to promote awareness of media bias. We experiment with an LLM-based approach to MBC generation, finding that retrieval-augmentation~\citep{lewis2020rag} improves performance. We carry out human evaluation, documenting 1) that humans find MBCs helpful when assessing retrieved documents, and 2) that LLMs augmented with MBCs in addition to their retrieved results give better answers to questions on controversial topics. We release our dataset and code at \url{https://github.com/MichSchli/MediaBackgroundChecks} (CC-BY-NC-4.0).

\section{Task Definition}
We seek to automatically \textit{find} and \textit{summarise} indicators of trustworthiness and tendency in order to facilitate source-critical reasoning by humans and NLP models. We refer to such summaries as \textit{media background checks} (MBC). A fact is considered a relevant indicator to a media background check for source X if it can shift the relative trust assigned \textit{to} source X for some consumer of content. Such facts include, but are not limited to, the founding, leadership, and organization of the source, the funding model of the source, the remit and target audience of the source, public stances taken by the source (e.g., endorsing a political campaign), discussion \textit{in} other media about the source, reliance \textit{by} other media on the source, and the factual history of the source (as indicated e.g. by fact-checks undergone by reputable fact-checking organisations).

We identify 42 common patterns of relevant facts, which we also use for evaluation -- see Appendix~\ref{appendix:atomic_fact_templates}. However, these do not fully cover the facts seen empirically in background checks from MB/FC. Real-world indicators also include rarer patterns, such as multi-hop facts (e.g. other sources owned by the same company), and topic-specific indicators, for example evidence that a source cannot be trusted specifically on issues related to healthcare, Québécois politics, or news about a particular advertiser. To understand common patterns, we further analyse 20 randomly chosen articles from MB/FC. The most common facts include known biases, funding model, remit, fact-checking history, and ownership (see Appendix~\ref{appendix:facts_in_mbcs}).

In order to build an MBC, a model must first recall or retrieve pertinent details for the source. These range from simple facts (e.g., whether the source is funded by a specific government) to more complex (e.g., whether a poor track record for factuality is evidenced). Necessary retrieval steps may differ depending on the source type -- websites have different indicators of credibility than print sources, for example~\citep{potter2013media}. Once the appropriate information has been retrieved, the model must then summarise this information.

\section{Dataset}
\label{section:dataset}

Media Bias / Fact Check\footnote{https://mediabiasfactcheck.com/} (MB/FC) is an independent, volunteer-run website that promotes awareness of media bias. While past datasets have used bias ratings from the site~\citep{baly-etal-2018-predicting}, the textual \textit{``detailed reports''} are an untapped resource. These are well-sources documents that summarise the factual history, bias, tendency, and potential conflicts of interest for media sources -- i.e., \textit{background checks}. We introduce a novel dataset based on these. We collected all reports listed through their search tool. This provided a total of 6,709 background checks. We cleaned the reports, and removed the bias and credibility ratings. Statistics for the dataset can be seen in Table~\ref{table:dataset_stats}, and the dataset itself can be downloaded at \url{https://github.com/MichSchli/MediaBackgroundChecks}.

\begin{table}[tb]
\centering
\begin{tabular}{l|lll}
\toprule
                  & Train & Dev   & Test  \\ \midrule
Background checks & 5209  & 500   & 1000  \\
\# Avg. Lines          & 17.1  & 17.2  & 17.0  \\
\# Avg. Tokens         & 305.1 & 302.2 & 303.2 \\ \bottomrule
\end{tabular}
\caption{Statistics for our dataset. Background checks were randomly divided among the three splits.}
\label{table:dataset_stats}
\end{table}

While MB/FC is an extensive resource, covering almost 7,000 media organisations, it is far from complete. In our human evaluation in Sections~\ref{section:human_evaluation}, out of 40 sources used, 29 are covered by MB/FC reports. However, of these 29, 9 are incomplete or lack crucial information. 
We do not intend to replace human-written background checks -- rather, we suggest that generated background checks can supplement MB/FC in these cases.



\section{Experiments}
\subsection{Models}
\label{section:models}

For a simple baseline, we prompt a large language model to generate an MBC given the name of a source. Our prompt can be seen in Appendix~\ref{appendix:initial_prompt}. We experiment with two models, one closed-source and one open-source: \textit{gpt-3.5-turbo-0125}\footnote{ChatGPT; see \url{https://platform.openai.com/docs/models/gpt-3-5-turbo}} and \textit{Llama 3 8B Instruct}\footnote{see \url{https://huggingface.co/meta-llama/Meta-Llama-3-8B-Instruct}}, the instruction-tuned version of Llama 3. The latter is small enough to reasonably run on a single consumer-grade GPU. 

For these initial models, we rely solely on the information stored in their weights. As such, their performance is representative of the level of source-critical reasoning that can be expected of current-generation LLMs, without any additional evidence provided. MBCs often incorporate information available on the internet -- the 6,709 background checks we analysed contained an average of 8.9 hyperlinks. 
To surface this information, we propose to rely on retrieval-augmented generation~\citep{lewis2020rag}. In addition to providing external memory at generation time, retrieved evidence can also be shown to the user to establish why the \textit{background check} should (or should not) be trusted. Showing where evidence originates from to establish trust is crucial in related tasks, such as automated fact-checking~\citep{guo-etal-2022-survey}. It also provides a rudimentary form of interpretability.

\begin{figure*}[ht]
\begin{tcolorbox}[colframe=black,colback=white,boxrule=0.2mm,arc=0mm,boxsep=1mm]
\textbf{Gold Example:} \\
History\\
Founded in 2005, Natural News is a conspiracy and pseudoscience website that routinely publishes false information. The founder is Mike Adams, who owns several Questionable websites such as News Target and Trump.news.\\

Funded by / Ownership\\
Natural News is owned by Mike Adams, who owns numerous other fake and or controversial websites. The website, like all of Mike Adams’s properties, is funded through online advertising.\\

Analysis / Bias\\
...
\vspace{0.5cm}
\hrule
\vspace{0.5cm}
\textbf{Generated Example (GPT-3.5-Turbo):} \\
Background check

1. Natural News is owned by Mike Adams, who is also known as the ``Health Ranger.''\\
2. The website is primarily funded through the sale of various health supplements, books, and other products advertised on the site.\\
3. Natural News is known for promoting alternative medicine, conspiracy theories, and pseudoscience. \\
4. The site has been criticized for its lack of credible sources and scientific evidence to support its claims.

\vspace{0.5cm}
\hrule
\vspace{0.5cm}
\textbf{Generated Example (GPT-3.5-Turbo + Google):} \\
**Background check on naturalnews.com:**
\\\\
1. **Ownership**: Founded by Mike Adams, known as the ``Health Ranger.''\\
2. **Funding**: Primarily funded through the sale of various health supplements and products on their website.\\
3. **Bias**: Strongly criticized for promoting pseudoscience, conspiracy theories, and misinformation, particularly regarding vaccines and alternative medicine.\\
4. **Fact-checks**: Natural News has a history of publishing misinformation and has been flagged by fact-checkers such as Snopes, Politifact, and FactCheck.org for spreading false information.\\
5. **Type of site**: Known for promoting alternative medicine, pseudoscience, disinformation, and far-right extremism.\\
6. **Country of Origin**: United States.\\
7. **Political leaning**: Natural News is considered a far-right website known for anti-vaccination conspiracy theories and fake news.
\end{tcolorbox}
\caption{Example background checks for \textit{Natural News}. The gold example is taken from the Media Bias / Fact Check website, while the generated example is produced by GPT-3.5 augmented with Google search as described in Section~\ref{section:models}. The gold example has been shortened, and the full version can be seen at \url{https://mediabiasfactcheck.com/natural-news/}.}
\label{fig:examples}
\end{figure*}

\begin{table*}[t]
\centering
\begin{tabular}{l|rr|rr}\toprule
& Fact Recall & Error Rate & METEOR & ROUGE-L \\ \midrule
GPT-3.5-Turbo & 22.7\% & \textbf{6.2\%} & 9.9\% & 12.5\% \\
GPT-3.5-Turbo + Google & \textbf{26.1\%} & 6.3\% & 12.6\% & 13.1\% \\
Llama 3 7b Instruct & 24.4\% & 10.4\% & 15.3\% & \textbf{14.4\%}\\
Llama 3 7b Instruct + Google & 25.1\% & 10.7\% & \textbf{15.5\%} & \textbf{14.4\%} \\\bottomrule
\end{tabular}
\caption{Performance of four different systems on our MB/FC dataset. We measure fact recall and error rate, as discussed in Section~\ref{section:evaluation}. These represent, respectively, how many facts from the gold background check were recalled by the model, and how many facts from the gold background check were \textit{contradicted} by the model. We also include two traditional summarisation metrics, METEOR and ROUGE-L.}
\label{table:performance}
\end{table*}

As the necessary retrieval steps are roughly similar across background checks, we have compiled a list of seven information-seeking search queries focusing on different aspects of a background check (see the list in Appendix~\ref{appendix:information_seeking_questions}). In initial experiments, we found this strategy to perform better than generating queries using our LLM. 

For each query, we gather relevant search results using the Google Search API\footnote{https://programmablesearchengine.google.com/}, selecting the top 30 retrieved documents. We exclude the Media Bias / Fact Check website and any website linking to it. To focus on the most salient retrieved information, we use a question-answering model to extract answers to a predefined question for each information-seeking query (also found in Appendix~\ref{appendix:information_seeking_questions}). Here, we employ DeBERTa v.3~\citep{he2021deberta} pretrained on SQuAD~\citep{rajpurkar-etal-2016-squad}. By this process we transform a Google search result into a question-answer pair.

For each question-answer pair, we then \textit{expand} the background check with the information contained within. This allows us to gradually incorporate information from multiple rounds of retrieval, without running into the token limit. To expand a background check with information from a specific question-answer pair, we use the prompt seen in Appendix~\ref{appendix:update_prompt} with each respective LLM.

\subsection{Evaluation}
\label{section:evaluation}

When evaluating media background checks, we are primarily interested in \textit{recalling} the information provided in the gold background check. 
Systems may find information beyond what is included in the gold example. Prioritising recall avoids penalising such additions.

We adopt a variant of FActScores~\citep{min-etal-2023-factscore} to evaluate models. We begin by breaking each gold MBC down into atomic facts. As MBCs cover relatively similar topics, we use a list of 42 templates to generate facts (see Appendix~\ref{appendix:atomic_fact_templates}). Each template contains contextually determined tokens, which we fill with details from the gold MBC using \textit{GPT-3.5-turbo}. During initial experiments, we found this strategy to perform better than fully generating facts. We subsequently verify whether each fact is entailed by the gold MBC. We experimented with several models, including a trained DeBERTa model~\citep{he2021deberta} and an open-source LLM
, but ultimately found the best-performing system to be \textit{GPT-3.5-turbo} with zero-shot chain of though~\citep{kojima2022step}. Our prompt can be found in  Appendix~\ref{appendix:atomic_fact_prompt}. To increase performance, we prompt the (probabilistic) model four times, and take as prediction as the majority-voted element among the runs. We keep atomic facts which are \textit{entailed} or \textit{contradicted} by the gold MBC.

To score a generated background check, we compute \textit{fact recall}, the percentage of atomic facts for which the gold MBC and the predicted MBC agree on entailment. We use the same entailment model
. We furthermore report the error rate, the percentage of atomic facts either are \textit{entailed} by one MBC and \textit{contradicted} by the other.

As a sanity check, we evaluate what performance the \textit{gold} background checks would yield on the development set; that is, using gold data both as prediction and reference. Ideally, the fact recall rate should be roughly 1, and the error rate roughly 0; but as entailment is noisy, this ideal is not reached. The computed fact recall and error rate are, respectively, 84.1\% and 3.4\%. We conclude that fact recall, while noisy, remains a useful measurement of how many gold facts are recalled.

We furthermore manually inspect evaluation for ten MBCs. We find that, on average, 59.8\% of the facts from the gold summary were accounted for in the generated atomic facts. The majority of the excluded facts were ``multi-hop facts'', which unfortunately our evaluation strategy does not account for -- such as which \textit{other} media companies are owned by the parent organization of the source reported on in the background check.

When the right atomic facts \textit{are} extracted, our entailment system works well. When evaluating generated MBCs for these ten sources, we agree with 93.9\% of entailment predictions made by our GPT-3.5 ensemble.

\subsection{Results}
\label{section:results}

Using this evaluation metric, we evaluate our four MBC generation models (see Table~\ref{table:performance}). In addition to our proposed metric, we also include two traditional measures: METEOR and ROUGE-L. Fact recall rates are low, highlighting the difficulty of the task. Nevertheless, we see clear performance improvements from retrieval-augmentation, both for GPT-3.5 and Llama 3. This supports our intuition that \textit{finding the right information} is a crucial barrier to source-critical reasoning in models.

To understand what causes the low recall, we manually analyse 10 randomly chosen examples (generated by GPT-3.5 with retrieval). We find the following omissions: seven missing entity mentions (e.g., who the editor is), three missing failed fact-checks, three missing mentions of editorial stance (e.g., right- or left-leaning), two missing funding sources, two historical events (e.g., an ownership change), and one missing mention of content being re-published from another site. Without retrieval, we see a further two missing  failed fact-checks, one missing entity mention, one missing funding source, and one missing mention of editorial stance. 

We furthermore see, in both cases, five different multi-hop facts omitted. However, as we discuss Section~\ref{section:evaluation}, our evaluation metric \textit{also} fails at catching those, so they are not responsible for the low recall. Overall, the biggest problem is missing entities -- future work could tailor retrieval especially to this scenario, for example by placing extra emphasis on retrieved “about”-sections.

Compared to GPT-3.5, Llama achieves high fact recall and METEOR/ROUGE-L scores, but also exhibits a high error rate. Exploring the data, we see significant differences in the lengths of MBCs -- the average GPT-3.5 generated MBC is 176 tokens, while for Llama the average has 254 tokens. The gold MBCs contain on average 477 tokens. This explains the discrepancy: Llama generates longer summaries with more facts, and so is more likely to state both correct and incorrect things about the knowledge source. Manually reviewing the generated data, we see one more difference: GPT-3.5 performs better at integrating retrieved information, explaining the higher fact recall in the retrieval-augmented setting.

\begin{table*}[htb]
\centering
\begin{tabular}{lccccccc}
\toprule
 & \multicolumn{2}{c}{with MBC} & \multicolumn{2}{c}{without MBC} & & \\
 & mean & SD & mean & SD & \textit{t} Statistic & \textit{p}-value \\
\midrule
Provision of Sufficient Information & 78.2\% & - & 70.9\% & - & 0.740 & 0.746 \\
Difficulty of Answering & 2.24 & 0.75 & 2.81 & 0.82 & -1.633 & 0.133 \\
Difficulty of Establishing Trust & 1.95 & 0.54 & 2.88 & 0.65 & -3.791* & 0.004 \\
\bottomrule
\end{tabular}
\caption{Results for part one of our human evaluation, estimating the helpfulness of generated MBCs when presented directly to users. Provision of sufficient information is annotated as a binary yes/no-question, while the difficulty of answering the question and the difficulty of establishing which sources are trustworthy are rated on five-point Likert scales. Results are analysed via student's \textit{t}-tests. * indicates significance at $p=0.01$.}
\label{table:survey_part_one}
\end{table*}

\subsection{Human Evaluation}
\label{section:human_evaluation}

We envision MBCs as being used in two settings: as assistive instruments for either \textit{humans} or \textit{models} having to make sense out of untrustworthy evidence. To demonstrate the potential, we conduct two experiments with human participants. For this purpose, we recruited 11 researchers working on automated fact-checking, hate speech analysis, LLMs, and related NLP tasks (ranging from PhD-students to assistant professors).

\subsubsection{QA with Untrustworthy Evidence}

We first create a small dataset of questions for which multiple, conflicting, and potentially untrustworthy evidence documents could reasonably be expected to surface. We compose this of ten questions about \textit{controversial subjects}, and ten questions about \textit{known misinformation}. \citet{wan2024evidence} recently released a small dataset of questions with conflicting evidence, including four examples of controversial political questions. We include the four from their dataset, and generate an addition six following their approach. For known misinformation, we manually extract ten questions by rephrasing claims from the development set of the AVeriTeC dataset~\citep{schlichtkrull2023averitec}, which contains labelled examples of fact-checked real-world misinformation. We choose randomly from \textit{refuted} claims in the dataset, focusing on claims originating from sources easily findable via a search engine (i.e., excluding claims from Twitter, Facebook, and other social media). 

For all twenty questions, we find two disagreeing evidence documents. We pick the first two search results that disagree when entering the question into Google. In all cases, these documents are found on the first page. These are documents which a searcher, whether human or algorithm, would easily come across. For the ten \textit{known misinformation} questions, the original misinformation source labelled in AVeriTeC appeared on the first page 6/10 times; for the rest, we found an alternative source supporting the misinformation.

We believe question answering is representative of how LLMs are envisioned to replace search; see e.g. \citet{metzler2021rethinking}. We recognise that, in actual use, LLMs are also used for other tasks, such as programming and software engineering-related functions. Answering questions about culture and geography is among the most frequent uses~\citep{zheng2024lmsys}, and for such questions, source reliability is certainly an issue. Further, answering more general questions is also a common use-case. As such, we believe that QA is 1) \textit{highly} representative of the vision large providers have of LLMs, and 2) representative of current use cases.

\subsubsection{Are MBCs helpful for humans?}
\label{section:human_exp_1}

\begin{table*}[htb]
\centering
\begin{tabular}{lcccccc}
\toprule
 & with MBC & without MBC & Equally Good & $\chi^2$ & $p$-value \\
\midrule
Preferred Answer & 165 & 26 & 29 & 57.02* & 0.000  \\
Better Understanding Provided & 69 & 56 & 95 & 60.22* & 0.001  \\
\bottomrule
\end{tabular}
\caption{Results for part two of our human evaluation, estimating the helpfulness of generated MBCs when provided along with retrieved results to GPT-4 in a question-answering task. Annotators directly state which answer they prefer, and which provides a better understanding of the topic. Results are analysed via a chi-square test. * indicates significance at $p=0.01$.}
\label{table:survey_part_two_preference}
\end{table*}

\begin{table*}[htb]
\centering
\begin{tabular}{lccccccc}
\toprule
 & \multicolumn{2}{c}{with MBC} & \multicolumn{2}{c}{without MBC} & & \\
 & mean & SD & mean & SD & \textit{t} Statistic & $p$-value \\
\midrule
Answer is Misleading & 1.57 & 0.35 & 2.58 & 0.63 & -5.634* & 0.000 \\
\bottomrule
\end{tabular}
\caption{Further results for part two of our human evaluation. Annotators evaluate how misleading LLM responses are on a five-point Likert scale. Results are analysed via a student's \textit{t}-test. * indicates significance at $p=0.01$.}
\label{table:survey_part_two_misleading}
\end{table*}

We first investigate whether MBCs are helpful for humans when encountering conflicting sources. We ask our annotators to answer ten questions from our dataset, given the two conflicting sources associated with that question. We randomly pick five questions based on controversial subjects, and five based on misinformation. For each question, we further randomly choose whether to show an accompanying background check. We use generated MBCs, simulating real-world use of our model. We use the best-performing model -- GPT-3.5-Turbo with retrieval.

After answering the question, each participant is asked to judge 1) whether they were given sufficient information to answer the question; 2) how difficult it was to answer the question; and 3) how difficult it was to decide which source to trust. Intuitively, if MBCs are helpful for humans when engaging with epistemic uncertainty, we should expect a lower cognitive load -- and thus for the task to feel less difficult~\citep{sweller2011cognitive} -- when an MBC is provided. For information sufficiency, we ask participants to give a binary answer. For task difficulty, we follow previous work and use a self-reported Likert scale as our measurement. We use a scale of 1 (very easy) to 5 (very difficult). An example page from our questionnaire can be seen in Appendix~\ref{appendix:survey_part_1}.

In Table~\ref{table:survey_part_one}, we report the mean answer given by our participants for information sufficiency, answer difficulty, and trust difficulty. We furthermore report the standard difference between participants for answer difficulty and trust difficulty. To analyse our results, we conduct a paired samples t-test comparing the responses of each participant with and without media background checks.

The cognitive load of deciding which documents to trust was \textit{much} lower when annotators were provided with a background check (1.95 versus 2.88, on average). The cognitive load of answering the question was also lower with a background check, although this result is not statistically significant. Similarly, our annotators more frequently reported that sufficient information had been provided when also given a background check, but this result was not statistically significant either. Post-hoc discussions with our participants were revealing: when not provided with a background check, some participants instead used the internet to find similar information for themselves. As our instructions included a link to each of the sources used, participants considered the information provided to be ``sufficient''; although answering the question required additional effort. For humans, automatically generated background checks thus \textit{quicken} a part of the meaning-producing process that is \textit{in all cases necessary}.

\subsubsection{Are MBCs helpful for NLP models?}
\label{section:human_exp_2}

Ultimately, our goal in developing MBCs was to introduce and test source-critical reasoning capabilities in NLP models, specifically LLMs. As such, we also seek to demonstrate the helpfulness of MBCs to retrieval-augmented LLMs. To do so, we again conduct a human experiment. 

We simulate a question-answering setting with a retrieval-augmented LLM. Given one of the remaining questions from our dataset, we assume that retrieval has returned one of the two disagreeing evidence documents. We ask GPT-4\footnote{gpt-4-turbo-2024-04-09} to answer the question based on the returned document. We repeat this with the other evidence document. For each source, we generate a background check using our best-performing model (GPT-3.5-Turbo with search). We generate a second version of the answer, including this background check in the instructions to GPT-4. Our prompts for the two cases can be seen in Appendix~\ref{appendix:qa_prompts}.

We show our participants the two answers to each of these 20 question-source pairs (in random order). We then ask them to determine 1) which answer they prefer; 2) which answer gives the best understanding of the topic; 3) for each answer, if they would feel misled if given that answer by an AI chatbot. For preference and understanding, annotators can pick the first answer, the second answer, or indicate that the two answers are equally good. For feeling misled we employ a five-point Likert scale ranging from 1 (no, not at all) to 5 (yes, very much). An example page from our questionnaire can be seen in Appendix~\ref{appendix:survey_part_2}.

In Table~\ref{table:survey_part_two_preference} we report how many times our participants indicated that they preferred an answer generated with or without an MBC, and how many times they indicated that answers with or without MBCs provided a better understanding of the topic. We conduct a chi-squared test to establish the significance of our results. The responses strongly indicate that participants preferred the answers generated with MBCs. Further, answers generated with MBCs provided a better understanding of the topic more often -- by a smaller margin, but still significantly so. Interestingly, if we analyse only the answers generated with (what we consider to be) trustworthy sources, the latter finding disappears while the former remains. We theorise that MBC-backed answers give a better understanding when sources are untrustworthy, but are preferred even for trustworthy sources as they help the user obtain confidence in their answer.

In addition to the pairwise comparison, we further investigated misleadingness. In Table~\ref{table:survey_part_two_misleading}, we report the mean answer given by our participants on how misled they would feel if given a generated answer by a chatbot, along with the standard deviation. We conduct a paired samples \textit{t}-test to analyse these results. As can be seen, answers generated with MBCs were on average rated \textit{significantly less misleading} than answers generated without MBCs. 

These findings support our primary hypotheses -- that search-augmented LLMs do not adequately account for the tendencies and biases of the sources they rely on, and that providing (even automatically generated) MBCs to models at inference time can alleviate this and enable automated source-critical reasoning.

\section{Related Work}

Closest to our work, \citet{baly-etal-2018-predicting, zhang-etal-2019-tanbih, baly-etal-2020-written} classifies the bias and factuality of sources based on data from Media Bias/Fact Check. However, they only predict the \textit{bias labels} for both (as given elsewhere on MB/FC), \textit{not} the detailed background checks we produce. Features include sample articles from the source, its Wikipedia page and Twitter account, and information about the web domain. \citet{hounsel2020identifying} proposed further web-domain-features such as host, domain, and certificates for websites when predicting trustworthiness.

Knowledge conflicts in retrieved evidence for LLMs is an active research area; see the survey by~\citet{xu2024knowledge}. Using their terminology, our proposal is concerned with \textit{inter-context conflicts}. The mentioned mitigation strategies, e.g. \citet{chen2023combating, vergho2024comparing}, follow the above line of reasoning: they use a trained model to detect and remove ``unfactual'' documents. This assumes a single source of truth (the training data or evidence database for the misinformation detector), which as we have argued does not align with best practises for human knowledge experts.

A rich literature exists -- referred to as \textit{fact-finding} -- proposing probabilistic models for computing the likelihood of claims and the trustworthiness of sources~\citep{yin2007truthfinder, dong2009integrating, pasternack-roth-2010-knowing, pasternack2013latent, Vydiswaran2011content, zhang-etal-2019-evidence}. The basis is the assumption by \citep{yin2007truthfinder} that \textit{``a web site is trustworthy if it provides many pieces of true information, and a piece of information is likely to be true if it is provided by many trustworthy web sites''}. \citet{yuan-etal-2020-early} proposed to predict the credibility of sources based on whether known misinformation spreaders share similar content. Along similar lines, \citet{wright-augenstein-2021-citeworth} introduced a model for citation-worthiness where citation-worthy papers feature in citation-worthy journals, and vice versa.

\citet{dong2009integrating} identified a \textit{copying problem} in fact-finding, where this strategy breaks down if many seemingly independent actors copy their positions from each other; this makes majority voting an unreliable heuristic for determining whether a piece of information is trustworthy. \citet{bala_learning_1998} identified a similar issue, where a ``royal family of knowers'' can cause communities to converge on false beliefs if their connectivity is much greater than the average epistemic agent. 

\citet{kaneko-etal-2009-mediatory, nakano-etal-2010-construction} proposed to develop ``credibility survey reports'' for search results. These are topic-specific documents that map the positions of different, contradicting search results with respect to \textit{one particular query} and to each other. Unlike our proposal, these do not represent the \textit{general} biases and tendencies of authors in, and so limit reasoning entirely to what can be done on the basis of one search query. \citet{shibuki-etal-2010-method} later developed a summarisation algorithm for this task.

The credibility and bias of individual documents has also been studied previously. For Wikipedia articles, \citet{zeng2006revision, adler2007wiki} attempted to predict trustworthiness based on revision history. \citep{nakov-etal-2017-trust} studies the credibility of statements on community QA forums based on linguistic signals in the individual posts; \citet{baly-etal-2020-detect} similarly used the surface forms of text to predict the credibility of news articles.

Finally, the concept of credibility has been studied extensively outside NLP, e.g. in information science. For an overview, the framework by \citet{rieh_credibility_2007} is an excellent resource. Similarly, source critical methods are widely studied, especially in history; we recommend \citet{howell2001reliable} for an introduction.

\section{Conclusion}
We have introduced the task of media background check generation, a novel task wherein models summarise information about media sources to enable critical analysis. We have furthermore presented a dataset of 6,709 examples collected from the Media Bias / Fact Check website. We have investigated several baselines for the task, including both open-source and closed-source LLMs. Our findings demonstrate that retrieval-augmentation can greatly improve performance, and interestingly that open-source models are very competitive on this difficult task. Finally, our human evaluation gives strong evidence both that media background checks are helpful for humans when evaluating media sources, and helpful for models when generating answers based on retrieved sources.

\section{Limitations}
Our paper proposes to establish trust by providing information about bias and tendency. In our models, that information comes either from model weights or retrieved documents. As such, that information is \textit{itself} potentially untrustworthy. Taken seriously, this prompts \textit{another} round of retrieval to establish the trustworthiness of the background check; and then another, to establish the trustworthiness of \textit{that} round -- it's turtles all the way down. Ultimately, our proposal cannot conclusively establish trust; only establish it insofar as the user already trusts some sources. As users have different requirements in terms of which sources they trust and how many ``levels'' of trust they might want to explore, one possible solution could be an interactive system allowing users to expand background checks in a desired direction, similar to recent proposals for summarisation~\citep{shapira-etal-2021-extending}.

When evaluating systems with access to search, we excluded the Media Bias / Fact Check website itself (and associated websites) from search results. Our intention was to make our evaluation setup as fair as possible. However, this introduces two limitations: 1) ``in the wild'' systems may perform \textit{better} than in our evaluations, as they do have access to background checks from MB/FC; and 2) we may still overestimate the relative performance of retrieval-augmented systems slightly, as we ultimately cannot make sure that no website which quotes MB / FC was included in search results. For the latter, manual analysis of 20 examples did not turn up any such websites.

\section{Ethical Considerations}
The dataset presented in this paper is based on Media Bias / Fact Check, a volunteer-run project. Background checks included here may themselves be biased on incomplete, as may background checks produced by models trained or evaluated on our dataset. Furthermore, the machine learning models and search engine used for our models contain well-known biases~\citep{noble2018algorithms, bender2021dangers}. Acting on trustworthiness estimates arrived at through biased means, including automatically produced ranking decisions for evidence retrieval, risks 
causing epistemic harm~\citep{schlichtkrull-etal-2023-intended}. The datasets and models described in this paper are not intended for and should not be used for truth-telling, e.g. for the
design of automated content moderation systems.

We did not take any steps to anonymise the data. The claims discussed in our dataset are based on publicly available data from a journalistic publication, and concern public figures and events -- references to these are important to document the history of a publication. If any person included in our dataset (e.g., the owner of a particular media source) requests it, we will remove that example from the dataset.

\section*{Acknowledgements}
We would like to thank Nedjma Ousidhoum and Rui Cao for their helpful comments, discussions, and feedback. We would also like to thank the anonymous reviewers for their questions and comments that helped us improve the paper. Finally, we would like to thank Dave Van Zandt and the Media Bias / Fact Check team for lending us their data, as well as for their pioneering work on the subject.

\bibliography{anthology,custom}

\begin{thebibliography}{54}
\providecommand{\natexlab}[1]{#1}

\bibitem[{Adler and de~Alfaro(2007)}]{adler2007wiki}
B.~Thomas Adler and Luca de~Alfaro. 2007.
\newblock \href {https://doi.org/10.1145/1242572.1242608} {A content-driven reputation system for the wikipedia}.
\newblock In \emph{Proceedings of the 16th International Conference on World Wide Web}, WWW '07, page 261–270, New York, NY, USA. Association for Computing Machinery.

\bibitem[{Bala and Goyal(1998)}]{bala_learning_1998}
Venkatesh Bala and Sanjeev Goyal. 1998.
\newblock \href {https://doi.org/10.1111/1467-937X.00059} {Learning from {Neighbours}}.
\newblock \emph{Review of Economic Studies}, 65(3):595--621.

\bibitem[{Baly et~al.(2020{\natexlab{a}})Baly, Da~San~Martino, Glass, and Nakov}]{baly-etal-2020-detect}
Ramy Baly, Giovanni Da~San~Martino, James Glass, and Preslav Nakov. 2020{\natexlab{a}}.
\newblock \href {https://doi.org/10.18653/v1/2020.emnlp-main.404} {We can detect your bias: Predicting the political ideology of news articles}.
\newblock In \emph{Proceedings of the 2020 Conference on Empirical Methods in Natural Language Processing (EMNLP)}, pages 4982--4991, Online. Association for Computational Linguistics.

\bibitem[{Baly et~al.(2018)Baly, Karadzhov, Alexandrov, Glass, and Nakov}]{baly-etal-2018-predicting}
Ramy Baly, Georgi Karadzhov, Dimitar Alexandrov, James Glass, and Preslav Nakov. 2018.
\newblock \href {https://doi.org/10.18653/v1/D18-1389} {Predicting factuality of reporting and bias of news media sources}.
\newblock In \emph{Proceedings of the 2018 Conference on Empirical Methods in Natural Language Processing}, pages 3528--3539, Brussels, Belgium. Association for Computational Linguistics.

\bibitem[{Baly et~al.(2020{\natexlab{b}})Baly, Karadzhov, An, Kwak, Dinkov, Ali, Glass, and Nakov}]{baly-etal-2020-written}
Ramy Baly, Georgi Karadzhov, Jisun An, Haewoon Kwak, Yoan Dinkov, Ahmed Ali, James Glass, and Preslav Nakov. 2020{\natexlab{b}}.
\newblock \href {https://doi.org/10.18653/v1/2020.acl-main.308} {What was written vs. who read it: News media profiling using text analysis and social media context}.
\newblock In \emph{Proceedings of the 58th Annual Meeting of the Association for Computational Linguistics}, pages 3364--3374, Online. Association for Computational Linguistics.

\bibitem[{Bender and Friedman(2018)}]{bender-friedman-2018-data}
Emily~M. Bender and Batya Friedman. 2018.
\newblock \href {https://doi.org/10.1162/tacl_a_00041} {Data statements for natural language processing: Toward mitigating system bias and enabling better science}.
\newblock \emph{Transactions of the Association for Computational Linguistics}, 6:587--604.

\bibitem[{Bender et~al.(2021)Bender, Gebru, McMillan-Major, and Shmitchell}]{bender2021dangers}
Emily~M Bender, Timnit Gebru, Angelina McMillan-Major, and Shmargaret Shmitchell. 2021.
\newblock On the dangers of stochastic parrots: Can language models be too big?\,\raisebox{-2pt}{\includegraphics[scale=0.11]{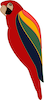}}.
\newblock In \emph{Proceedings of the 2021 ACM Conference on Fairness, Accountability, and Transparency}, pages 610--623.

\bibitem[{Bu\c{c}inca et~al.(2021)Bu\c{c}inca, Malaya, and Gajos}]{overreliance}
Zana Bu\c{c}inca, Maja~Barbara Malaya, and Krzysztof~Z. Gajos. 2021.
\newblock \href {https://doi.org/10.1145/3449287} {To trust or to think: Cognitive forcing functions can reduce overreliance on ai in ai-assisted decision-making}.
\newblock \emph{Proc. ACM Hum.-Comput. Interact.}, 5(CSCW1).

\bibitem[{Chen and Shu(2023)}]{chen2023combating}
Canyu Chen and Kai Shu. 2023.
\newblock Combating misinformation in the age of llms: Opportunities and challenges.
\newblock \emph{arXiv preprint arXiv:2311.05656}.

\bibitem[{Dong et~al.(2009)Dong, Berti-Equille, and Srivastava}]{dong2009integrating}
Xin~Luna Dong, Laure Berti-Equille, and Divesh Srivastava. 2009.
\newblock \href {https://doi.org/10.14778/1687627.1687690} {Integrating conflicting data: the role of source dependence}.
\newblock \emph{Proc. VLDB Endow.}, 2(1):550–561.

\bibitem[{Godler and Reich(2015)}]{Godler2015}
Yigal Godler and Zvi Reich. 2015.
\newblock \href {https://doi.org/10.1177/1464884915620268} {Journalistic evidence: Cross-verification as a constituent of mediated knowledge}.
\newblock \emph{Journalism: Theory, Practice \& Criticism}, 18(5):558--574.

\bibitem[{Guo et~al.(2022)Guo, Schlichtkrull, and Vlachos}]{guo-etal-2022-survey}
Zhijiang Guo, Michael Schlichtkrull, and Andreas Vlachos. 2022.
\newblock \href {https://doi.org/10.1162/tacl_a_00454} {A survey on automated fact-checking}.
\newblock \emph{Transactions of the Association for Computational Linguistics}, 10:178--206.

\bibitem[{He et~al.(2021)He, Liu, Gao, and Chen}]{he2021deberta}
Pengcheng He, Xiaodong Liu, Jianfeng Gao, and Weizhu Chen. 2021.
\newblock \href {https://openreview.net/forum?id=XPZIaotutsD} {Deberta: Decoding-enhanced bert with disentangled attention}.
\newblock In \emph{International Conference on Learning Representations}.

\bibitem[{Hounsel et~al.(2020)Hounsel, Holland, Kaiser, Borgolte, Feamster, and Mayer}]{hounsel2020identifying}
Austin Hounsel, Jordan Holland, Ben Kaiser, Kevin Borgolte, Nick Feamster, and Jonathan Mayer. 2020.
\newblock Identifying disinformation websites using infrastructure features.
\newblock In \emph{10th USENIX Workshop on Free and Open Communications on the Internet (FOCI 20)}.

\bibitem[{Howell and Prevenier(2001)}]{howell2001reliable}
Martha~C. Howell and William Prevenier. 2001.
\newblock \href {https://books.google.co.uk/books?id=wSqgwOZPjJ4C} {\emph{From Reliable Sources: An Introduction to Historical Methods}}.
\newblock Cornell paperbacks. Cornell University Press.

\bibitem[{Imperial et~al.(2024)Imperial, Forey, and Madabushi}]{imperial2024standardize}
J.~M. Imperial, G.~Forey, and H.~T. Madabushi. 2024.
\newblock \href {https://doi.org/10.48550/arXiv.2402.12593} {Standardize: Aligning language models with expert-defined standards for content generation}.
\newblock \emph{arXiv preprint arXiv:2402.12593}.

\bibitem[{{International Fact-Checking Network}(2016)}]{IFCN2016}
{International Fact-Checking Network}. 2016.
\newblock \href {https://ifcncodeofprinciples.poynter.org/} {Code of principles}.
\newblock Poynter Institute.

\bibitem[{Kaneko et~al.(2009)Kaneko, Shibuki, Nakano, Miyazaki, Ishioroshi, and Mori}]{kaneko-etal-2009-mediatory}
Koichi Kaneko, Hideyuki Shibuki, Masahiro Nakano, Rintaro Miyazaki, Madoka Ishioroshi, and Tatsunori Mori. 2009.
\newblock \href {https://aclanthology.org/Y09-1026} {Mediatory summary generation: Summary-passage extraction for information credibility on the web}.
\newblock In \emph{Proceedings of the 23rd Pacific Asia Conference on Language, Information and Computation, Volume 1}, pages 240--249, Hong Kong. City University of Hong Kong.

\bibitem[{Kojima et~al.(2022)Kojima, Gu, Reid, Matsuo, and Iwasawa}]{kojima2022step}
Takeshi Kojima, Shixiang~(Shane) Gu, Machel Reid, Yutaka Matsuo, and Yusuke Iwasawa. 2022.
\newblock \href {https://proceedings.neurips.cc/paper_files/paper/2022/file/8bb0d291acd4acf06ef112099c16f326-Paper-Conference.pdf} {Large language models are zero-shot reasoners}.
\newblock In \emph{Advances in Neural Information Processing Systems}, volume~35, pages 22199--22213. Curran Associates, Inc.

\bibitem[{Lewis et~al.(2020)Lewis, Perez, Piktus, Petroni, Karpukhin, Goyal, K\"{u}ttler, Lewis, Yih, Rockt\"{a}schel, Riedel, and Kiela}]{lewis2020rag}
Patrick Lewis, Ethan Perez, Aleksandra Piktus, Fabio Petroni, Vladimir Karpukhin, Naman Goyal, Heinrich K\"{u}ttler, Mike Lewis, Wen-tau Yih, Tim Rockt\"{a}schel, Sebastian Riedel, and Douwe Kiela. 2020.
\newblock Retrieval-augmented generation for knowledge-intensive nlp tasks.
\newblock In \emph{Proceedings of the 34th International Conference on Neural Information Processing Systems}, NIPS '20, Red Hook, NY, USA. Curran Associates Inc.

\bibitem[{Machill and Beiler(2009)}]{machill2009importance}
Marcel Machill and Markus Beiler. 2009.
\newblock \href {https://doi.org/10.1080/14616700802337768} {The importance of the internet for journalistic research: {A} multi-method study of the research performed by journalists working for daily newspapers, radio, television and online}.
\newblock \emph{Journalism Studies}, 10(2):178--203.

\bibitem[{Metzler et~al.(2021)Metzler, Tay, Bahri, and Najork}]{metzler2021rethinking}
Donald Metzler, Yi~Tay, Dara Bahri, and Marc Najork. 2021.
\newblock \href {https://doi.org/10.1145/3476415.3476428} {Rethinking search: making domain experts out of dilettantes}.
\newblock \emph{ACM SIGIR Forum}, 55:1--27.

\bibitem[{Min et~al.(2023)Min, Krishna, Lyu, Lewis, Yih, Koh, Iyyer, Zettlemoyer, and Hajishirzi}]{min-etal-2023-factscore}
Sewon Min, Kalpesh Krishna, Xinxi Lyu, Mike Lewis, Wen-tau Yih, Pang Koh, Mohit Iyyer, Luke Zettlemoyer, and Hannaneh Hajishirzi. 2023.
\newblock \href {https://doi.org/10.18653/v1/2023.emnlp-main.741} {{FA}ct{S}core: Fine-grained atomic evaluation of factual precision in long form text generation}.
\newblock In \emph{Proceedings of the 2023 Conference on Empirical Methods in Natural Language Processing}, pages 12076--12100, Singapore. Association for Computational Linguistics.

\bibitem[{Nakano et~al.(2010)Nakano, Shibuki, Miyazaki, Ishioroshi, Kaneko, and Mori}]{nakano-etal-2010-construction}
Masahiro Nakano, Hideyuki Shibuki, Rintaro Miyazaki, Madoka Ishioroshi, Koichi Kaneko, and Tatsunori Mori. 2010.
\newblock \href {http://www.lrec-conf.org/proceedings/lrec2010/pdf/135_Paper.pdf} {Construction of text summarization corpus for the credibility of information on the web}.
\newblock In \emph{Proceedings of the Seventh International Conference on Language Resources and Evaluation ({LREC}'10)}, Valletta, Malta. European Language Resources Association (ELRA).

\bibitem[{Nakov et~al.(2017)Nakov, Mihaylova, M{\`a}rquez, Shiroya, and Koychev}]{nakov-etal-2017-trust}
Preslav Nakov, Tsvetomila Mihaylova, Llu{\'\i}s M{\`a}rquez, Yashkumar Shiroya, and Ivan Koychev. 2017.
\newblock \href {https://doi.org/10.26615/978-954-452-049-6_072} {Do not trust the trolls: Predicting credibility in community question answering forums}.
\newblock In \emph{Proceedings of the International Conference Recent Advances in Natural Language Processing, {RANLP} 2017}, pages 551--560, Varna, Bulgaria. INCOMA Ltd.

\bibitem[{Noble(2018)}]{noble2018algorithms}
Safiya~Umoja Noble. 2018.
\newblock Algorithms of oppression.
\newblock In \emph{Algorithms of Oppression}. New York University Press.

\bibitem[{Pasternack and Roth(2010)}]{pasternack-roth-2010-knowing}
Jeff Pasternack and Dan Roth. 2010.
\newblock \href {https://aclanthology.org/C10-1099} {Knowing what to believe (when you already know something)}.
\newblock In \emph{Proceedings of the 23rd International Conference on Computational Linguistics (Coling 2010)}, pages 877--885, Beijing, China. Coling 2010 Organizing Committee.

\bibitem[{Pasternack and Roth(2013)}]{pasternack2013latent}
Jeff Pasternack and Dan Roth. 2013.
\newblock \href {https://doi.org/10.1145/2488388.2488476} {Latent credibility analysis}.
\newblock In \emph{Proceedings of the 22nd International Conference on World Wide Web}, WWW '13, page 1009–1020, New York, NY, USA. Association for Computing Machinery.

\bibitem[{Peskoff and Stewart(2023)}]{peskoff-stewart-2023-credible}
Denis Peskoff and Brandon Stewart. 2023.
\newblock \href {https://doi.org/10.18653/v1/2023.acl-short.37} {Credible without credit: Domain experts assess generative language models}.
\newblock In \emph{Proceedings of the 61st Annual Meeting of the Association for Computational Linguistics (Volume 2: Short Papers)}, pages 427--438, Toronto, Canada. Association for Computational Linguistics.

\bibitem[{Potter(2013)}]{potter2013media}
William~James. Potter. 2013.
\newblock \href {https://books.google.co.uk/books?id=TtMwJ85nK2UC} {\emph{Media Literacy}}.
\newblock SAGE Publications.

\bibitem[{Rajpurkar et~al.(2016)Rajpurkar, Zhang, Lopyrev, and Liang}]{rajpurkar-etal-2016-squad}
Pranav Rajpurkar, Jian Zhang, Konstantin Lopyrev, and Percy Liang. 2016.
\newblock \href {https://doi.org/10.18653/v1/D16-1264} {{SQ}u{AD}: 100,000+ questions for machine comprehension of text}.
\newblock In \emph{Proceedings of the 2016 Conference on Empirical Methods in Natural Language Processing}, pages 2383--2392, Austin, Texas. Association for Computational Linguistics.

\bibitem[{Rieh and Danielson(2007)}]{rieh_credibility_2007}
Soo~Young Rieh and David~R. Danielson. 2007.
\newblock \href {https://doi.org/10.1002/aris.2007.1440410114} {Credibility: {A} multidisciplinary framework}.
\newblock \emph{Annual Review of Information Science and Technology}, 41(1):307--364.

\bibitem[{Schlichtkrull et~al.(2023{\natexlab{a}})Schlichtkrull, Ousidhoum, and Vlachos}]{schlichtkrull-etal-2023-intended}
Michael Schlichtkrull, Nedjma Ousidhoum, and Andreas Vlachos. 2023{\natexlab{a}}.
\newblock \href {https://doi.org/10.18653/v1/2023.findings-emnlp.577} {The intended uses of automated fact-checking artefacts: Why, how and who}.
\newblock In \emph{Findings of the Association for Computational Linguistics: EMNLP 2023}, pages 8618--8642, Singapore. Association for Computational Linguistics.

\bibitem[{Schlichtkrull et~al.(2023{\natexlab{b}})Schlichtkrull, Guo, and Vlachos}]{schlichtkrull2023averitec}
Michael~Sejr Schlichtkrull, Zhijiang Guo, and Andreas Vlachos. 2023{\natexlab{b}}.
\newblock \href {https://openreview.net/forum?id=fKzSz0oyaI} {{AV}eri{T}e{C}: A dataset for real-world claim verification with evidence from the web}.
\newblock In \emph{Thirty-seventh Conference on Neural Information Processing Systems Datasets and Benchmarks Track}.

\bibitem[{Shah and Bender(2022)}]{shah2022situtating}
Chirag Shah and Emily~M. Bender. 2022.
\newblock \href {https://doi.org/10.1145/3498366.3505816} {Situating search}.
\newblock In \emph{Proceedings of the 2022 Conference on Human Information Interaction and Retrieval}, CHIIR '22, page 221–232, New York, NY, USA. Association for Computing Machinery.

\bibitem[{Shapira et~al.(2021)Shapira, Pasunuru, Ronen, Bansal, Amsterdamer, and Dagan}]{shapira-etal-2021-extending}
Ori Shapira, Ramakanth Pasunuru, Hadar Ronen, Mohit Bansal, Yael Amsterdamer, and Ido Dagan. 2021.
\newblock \href {https://doi.org/10.18653/v1/2021.naacl-main.54} {Extending multi-document summarization evaluation to the interactive setting}.
\newblock In \emph{Proceedings of the 2021 Conference of the North American Chapter of the Association for Computational Linguistics: Human Language Technologies}, pages 657--677, Online. Association for Computational Linguistics.

\bibitem[{Shibuki et~al.(2010)Shibuki, Nagai, Nakano, Miyazaki, Ishioroshi, and Mori}]{shibuki-etal-2010-method}
Hideyuki Shibuki, Takahiro Nagai, Masahiro Nakano, Rintaro Miyazaki, Madoka Ishioroshi, and Tatsunori Mori. 2010.
\newblock \href {https://aclanthology.org/C10-2131} {A method for automatically generating a mediatory summary to verify credibility of information on the web}.
\newblock In \emph{Coling 2010: Posters}, pages 1140--1148, Beijing, China. Coling 2010 Organizing Committee.

\bibitem[{Simpson(2013)}]{simpson2013evaluating}
Thomas~W Simpson. 2013.
\newblock Evaluating google as an epistemic tool.
\newblock \emph{Philosophical Engineering: Toward a Philosophy of the Web}, pages 97--115.

\bibitem[{Smith and Rieh(2019)}]{smith2019knowledge}
Catherine~L. Smith and Soo~Young Rieh. 2019.
\newblock \href {https://doi.org/10.1145/3295750.3298940} {Knowledge-context in search systems: Toward information-literate actions}.
\newblock In \emph{Proceedings of the 2019 Conference on Human Information Interaction and Retrieval}, CHIIR '19, page 55–62, New York, NY, USA. Association for Computing Machinery.

\bibitem[{Steensen(2019)}]{steensen_journalisms_2019}
Steen Steensen. 2019.
\newblock \href {https://doi.org/10.1177/1464884918809271} {Journalism’s epistemic crisis and its solution: {Disinformation}, datafication and source criticism}.
\newblock \emph{Journalism}, 20(1):185--189.
\newblock Publisher: SAGE Publications.

\bibitem[{Sweller et~al.(2011)Sweller, Ayres, and Kalyuga}]{sweller2011cognitive}
J.~Sweller, P.~Ayres, and S.~Kalyuga. 2011.
\newblock \href {https://books.google.co.uk/books?id=sSAwbd8qOAAC} {\emph{Cognitive Load Theory}}.
\newblock Explorations in the Learning Sciences, Instructional Systems and Performance Technologies. Springer New York.

\bibitem[{Thorne et~al.(2018)Thorne, Vlachos, Christodoulopoulos, and Mittal}]{thorne-etal-2018-fever}
James Thorne, Andreas Vlachos, Christos Christodoulopoulos, and Arpit Mittal. 2018.
\newblock \href {https://doi.org/10.18653/v1/N18-1074} {{FEVER}: a large-scale dataset for fact extraction and {VER}ification}.
\newblock In \emph{Proceedings of the 2018 Conference of the North {A}merican Chapter of the Association for Computational Linguistics: Human Language Technologies, Volume 1 (Long Papers)}, pages 809--819, New Orleans, Louisiana. Association for Computational Linguistics.

\bibitem[{Vergho et~al.(2024)Vergho, Godbout, Rabbany, and Pelrine}]{vergho2024comparing}
Tyler Vergho, Jean-Francois Godbout, Reihaneh Rabbany, and Kellin Pelrine. 2024.
\newblock \href {https://arxiv.org/abs/2401.06920} {Comparing gpt-4 and open-source language models in misinformation mitigation}.

\bibitem[{Vydiswaran et~al.(2011)Vydiswaran, Zhai, and Roth}]{Vydiswaran2011content}
V.G.~Vinod Vydiswaran, ChengXiang Zhai, and Dan Roth. 2011.
\newblock \href {https://doi.org/10.1145/2020408.2020567} {Content-driven trust propagation framework}.
\newblock In \emph{Proceedings of the 17th ACM SIGKDD International Conference on Knowledge Discovery and Data Mining}, KDD '11, page 974–982, New York, NY, USA. Association for Computing Machinery.

\bibitem[{Wadden et~al.(2020)Wadden, Lin, Lo, Wang, van Zuylen, Cohan, and Hajishirzi}]{wadden-etal-2020-fact}
David Wadden, Shanchuan Lin, Kyle Lo, Lucy~Lu Wang, Madeleine van Zuylen, Arman Cohan, and Hannaneh Hajishirzi. 2020.
\newblock \href {https://doi.org/10.18653/v1/2020.emnlp-main.609} {Fact or fiction: Verifying scientific claims}.
\newblock In \emph{Proceedings of the 2020 Conference on Empirical Methods in Natural Language Processing (EMNLP)}, pages 7534--7550, Online. Association for Computational Linguistics.

\bibitem[{Wan et~al.(2024)Wan, Wallace, and Klein}]{wan2024evidence}
Alexander Wan, Eric Wallace, and Dan Klein. 2024.
\newblock \href {https://arxiv.org/abs/2402.11782} {What evidence do language models find convincing?}
\newblock \emph{Preprint}, arXiv:2402.11782.

\bibitem[{Wright and Augenstein(2021)}]{wright-augenstein-2021-citeworth}
Dustin Wright and Isabelle Augenstein. 2021.
\newblock \href {https://doi.org/10.18653/v1/2021.findings-acl.157} {{C}ite{W}orth: Cite-worthiness detection for improved scientific document understanding}.
\newblock In \emph{Findings of the Association for Computational Linguistics: ACL-IJCNLP 2021}, pages 1796--1807, Online. Association for Computational Linguistics.

\bibitem[{Xu et~al.(2024)Xu, Qi, Guo, Wang, Wang, Zhang, and Xu}]{xu2024knowledge}
Rongwu Xu, Zehan Qi, Zhijiang Guo, Cunxiang Wang, Hongru Wang, Yue Zhang, and Wei Xu. 2024.
\newblock \href {https://arxiv.org/abs/2403.08319} {Knowledge conflicts for llms: A survey}.
\newblock \emph{Preprint}, arXiv:2403.08319.

\bibitem[{Yin et~al.(2007)Yin, Han, and Yu}]{yin2007truthfinder}
Xiaoxin Yin, Jiawei Han, and Philip~S. Yu. 2007.
\newblock \href {https://doi.org/10.1145/1281192.1281309} {Truth discovery with multiple conflicting information providers on the web}.
\newblock In \emph{Proceedings of the 13th ACM SIGKDD International Conference on Knowledge Discovery and Data Mining}, KDD '07, page 1048–1052, New York, NY, USA. Association for Computing Machinery.

\bibitem[{Yuan et~al.(2020)Yuan, Ma, Zhou, Han, and Hu}]{yuan-etal-2020-early}
Chunyuan Yuan, Qianwen Ma, Wei Zhou, Jizhong Han, and Songlin Hu. 2020.
\newblock \href {https://doi.org/10.18653/v1/2020.coling-main.475} {Early detection of fake news by utilizing the credibility of news, publishers, and users based on weakly supervised learning}.
\newblock In \emph{Proceedings of the 28th International Conference on Computational Linguistics}, pages 5444--5454, Barcelona, Spain (Online). International Committee on Computational Linguistics.

\bibitem[{Zeng et~al.(2006)Zeng, Alhossaini, Ding, Fikes, and McGuinness}]{zeng2006revision}
Honglei Zeng, Maher~A. Alhossaini, Li~Ding, Richard Fikes, and Deborah~L. McGuinness. 2006.
\newblock \href {https://doi.org/10.1145/1501434.1501445} {Computing trust from revision history}.
\newblock In \emph{Proceedings of the 2006 International Conference on Privacy, Security and Trust: Bridge the Gap Between PST Technologies and Business Services}, PST '06, New York, NY, USA. Association for Computing Machinery.

\bibitem[{Zhang et~al.(2019{\natexlab{a}})Zhang, Ives, and Roth}]{zhang-etal-2019-evidence}
Yi~Zhang, Zachary Ives, and Dan Roth. 2019{\natexlab{a}}.
\newblock \href {https://doi.org/10.18653/v1/P19-1040} {Evidence-based trustworthiness}.
\newblock In \emph{Proceedings of the 57th Annual Meeting of the Association for Computational Linguistics}, pages 413--423, Florence, Italy. Association for Computational Linguistics.

\bibitem[{Zhang et~al.(2019{\natexlab{b}})Zhang, Da~San~Martino, Barr{\'o}n-Cede{\~n}o, Romeo, An, Kwak, Staykovski, Jaradat, Karadzhov, Baly, Darwish, Glass, and Nakov}]{zhang-etal-2019-tanbih}
Yifan Zhang, Giovanni Da~San~Martino, Alberto Barr{\'o}n-Cede{\~n}o, Salvatore Romeo, Jisun An, Haewoon Kwak, Todor Staykovski, Israa Jaradat, Georgi Karadzhov, Ramy Baly, Kareem Darwish, James Glass, and Preslav Nakov. 2019{\natexlab{b}}.
\newblock \href {https://doi.org/10.18653/v1/D19-3038} {{T}anbih: Get to know what you are reading}.
\newblock In \emph{Proceedings of the 2019 Conference on Empirical Methods in Natural Language Processing and the 9th International Joint Conference on Natural Language Processing (EMNLP-IJCNLP): System Demonstrations}, pages 223--228, Hong Kong, China. Association for Computational Linguistics.

\bibitem[{Zheng et~al.(2024)Zheng, Chiang, Sheng, Li, Zhuang, Wu, Zhuang, Li, Lin, Xing, Gonzalez, Stoica, and Zhang}]{zheng2024lmsys}
Lianmin Zheng, Wei-Lin Chiang, Ying Sheng, Tianle Li, Siyuan Zhuang, Zhanghao Wu, Yonghao Zhuang, Zhuohan Li, Zi~Lin, Eric~P. Xing, Joseph~E. Gonzalez, Ion Stoica, and Hao Zhang. 2024.
\newblock \href {https://openreview.net/forum?id=Fg4vMW8NEQ} {Lmsys-chat-1m: A large-scale real-world llm conversation dataset}.
\newblock In \emph{International Conference on Learning Representations (ICLR)}.

\end{thebibliography}

\appendix

\section{Computational Resources}

Our experiments with opensource models were carried out using a single NVidia A100 GPU.

\section{Prompts}
\subsection{Initial Prompt}
\label{appendix:initial_prompt}

Our baseline consists of a simple prompt to an LLM, querying the model to generate an MBC based on the name of a media source. We use the same prompt for ChatGPT and Llama 3. The prompt can be seen in Figure~\ref{figure:initial_prompts}. When generating an initial background check to use as a starting point for our retrieval-augmented models, we use this prompt as well.

\begin{figure}[h]
    \centering
    \begin{tcolorbox}
{\fontfamily{qpl}\selectfont
\textbf{System message}
You are InfoHuntGPT, a world-class AI assistant used by journalists to quickly build knowledge of new sources.}\\

{\fontfamily{qpl}\selectfont
\textbf{User message}
Build a background check for the news source ``\{\textit{source name}\}''. Write down everything you know about them, e.g. who funds them, how they make money, if they have any particular bias. Make an ITEMIZED LIST. Be brief, and if you don't know something, just leave it out. If you are aware that they have failed any fact-checks, mention which. Begin your response with ``**Background check**''.}

\end{tcolorbox}
    \caption{Prompt used for ChatGPT and Llama 3 when generating MBCs with no supporting retrieved evidence. This prompt is also used to generate the \textit{initial} MBC which is later updated in the retrieved-evidence setting.}
    \label{figure:initial_prompts}
\end{figure}

\subsection{Update Prompt}
\label{appendix:update_prompt}
For our retrieval-augmented models, we use the prompt seen in Figure~\ref{figure:update_prompt} to incorporate new information from Google Search into a background check.

\begin{figure}[t]
    \centering
    \begin{tcolorbox}
{\fontfamily{qpl}\selectfont
\textbf{System message}
You are InfoHuntGPT, a world-class AI assistant used by journalists to quickly build knowledge of new sources.}\\

{\fontfamily{qpl}\selectfont
\textbf{User message}
Build a background check for the news source ``\{\textit{source name}\}''. Write down everything you know about them, e.g. who funds them, how they make money, if they have any particular bias. Make an ITEMIZED LIST. Be brief, and if you don't know something, just leave it out. If you are aware that they have failed any fact-checks, mention which. Begin your response with ``**Background check**''.}\\

{\fontfamily{qpl}\selectfont
\textbf{Assistant message}
\{\textit{Previous background check}\}}\\

{\fontfamily{qpl}\selectfont
\textbf{User message}
Google search has revealed some new information:\\\\
\{\textit{Question-answer pairs}\}\\\\
Update your background check for ``\{\textit{source name}\}'' using the new information. Do NOT delete any information, but make ADDITIONS where necessary, using the new information. Most likely, you will just need to add an extra item to the itemized list you previously created. Make minimal edits, and only incorporate what is relevant. Begin your response with ``**Background check**''.}

\end{tcolorbox}
    \caption{Prompt used for ChatGPT and Llama 3 when updating an MBC with retrieved information. The retrieved information is input to the prompt in the form of question-answer pairs, following the methodology described in Section~\ref{section:models}.}
    \label{figure:update_prompt}
\end{figure}

\subsection{Entailment Prompt}
\label{appendix:atomic_fact_prompt}

When evaluating MBCs via FActScores, we compute entailment using the prompt seen in Figure~\ref{figure:entailment_prompt}.

\begin{figure}[h]
    \centering
    \begin{tcolorbox}
{\fontfamily{qpl}\selectfont
\textbf{System message}
You are FactCheckGPT, a world-class tool used by journalists to discover problems in their writings. Users give you text, and check whether facts are true given the text. You ALWAYS answer either TRUE, FALSE, or NOT ENOUGH EVIDENCE.}\\

{\fontfamily{qpl}\selectfont
\textbf{User message}
You will be given a snippet written as part of a source criticism exercise, and a claim. Your task is to determine whether the claim is true based ONLY on the text. Do NOT use any other knowledge source\\\\
The claim is: ``\{\textit{hypothesis}\}''.\\
The text follows below:\\``\{\textit{premise}\}''.\\\\
\{\textit{hypothesis}\}? Thinking step by step, answer either TRUE, FALSE, or NOT ENOUGH EVIDENCE, capitalizing all letters. Explain your reasoning FIRST, and after that output either TRUE, FALSE, or NOT ENOUGH EVIDENCE.
}

\end{tcolorbox}
    \caption{Prompt used for our LLMs when computing textual entailment. We use a probabilistic model, and retain the majority prediction over four votes.}
    \label{figure:entailment_prompt}
\end{figure}

\subsection{Question-Answering Prompt}
\label{appendix:qa_prompts}

For our human evaluation in Section~\ref{section:human_exp_2}, we simulate retrieval-augmented question-answering. Given a source document and possibly an MBC, we answer questions using the prompts in Figures~\ref{figure:qa_prompt_without_mbc} and~\ref{figure:qa_prompt_with_mbc}.

\begin{figure}[h]
    \centering
    \begin{tcolorbox}
{\fontfamily{qpl}\selectfont
\textbf{System message}
You are an expert journalist. You will be given some evidence, and a question. Using the provided evidence, answer the question.}\\

{\fontfamily{qpl}\selectfont
\textbf{User message}
Using the provided evidence, answer the following: ``\{\textit{question}\}''.\\\\
The following evidence is provided: ``\{\textit{source document}\}\\\\
This information comes from ``\{\textit{domain name of source document}\}''.\\\\
Your answer should be short and concise, five sentences at most. Include an explanation of why the user should or should not trust the source.
}

\end{tcolorbox}
    \caption{Prompt used for answering questions using GPT-4 when not providing an MBC.}
    \label{figure:qa_prompt_without_mbc}
\end{figure}

\begin{figure}[h]
    \centering
    \begin{tcolorbox}
{\fontfamily{qpl}\selectfont
\textbf{System message}
You are an expert journalist. You will be given some evidence, and a question. Using the provided evidence, answer the question.}\\

{\fontfamily{qpl}\selectfont
\textbf{User message}
Using the provided evidence, answer the following: ``\{\textit{question}\}''.\\\\
The following evidence is provided: ``\{\textit{source document}\}\\\\
This information comes from ``\{\textit{domain name of source document}\}''.\\\\
The following information about the source may be relevant:\\
``\{\textit{background check}\}''\\\\
Your answer should be short and concise, five sentences at most. Include an explanation of why the user should or should not trust the source.
}

\end{tcolorbox}
    \caption{Prompt used for answering questions using GPT-4 when providing an MBC.}
    \label{figure:qa_prompt_with_mbc}
\end{figure}

\section{Fact Generation}
\label{appendix:atomic_fact_templates}
When evaluating background checks, we use a variant of FActScore~\citep{min-etal-2023-factscore}. The atomic facts we use are generated following the templates seen in Figure~\ref{figure:atomic_fact_templates}. The initial questions are decorated with information from the gold background check using the prompt seen in Figure~\ref{figure:atomic_fact_generation_prompt}.

\begin{figure}[t]
    \centering
    \begin{tcolorbox}
{\fontfamily{qpl}\selectfont
\textbf{System message}
You are InfoHuntGPT, a world-class AI assistant used by journalists to quickly build knowledge of new sources.}\\

{\fontfamily{qpl}\selectfont
\textbf{User message}
You will be given a snippet written as part of a source criticism exercise, and a fill-in-the-blank question (blanks represented by \textunderscore). Your task is to fill in the blanks in the sentence, adding no additional information or wording. JUST replace the \textunderscore character.\\\\
The question is:\\ 
\{\textit{template}\}\\\\
The text follows below:\\
\{\textit{gold background check}\}
\\\\
Fill in the blanks in the question, adding no additional information or wording. JUST replace the \textunderscore character, and output ONLY the question with the blank filled in. No yapping.
}

\end{tcolorbox}
    \caption{Prompt used for ChatGPT when decorating an atomic fact template with information from a background check.}
    \label{figure:atomic_fact_generation_prompt}
\end{figure}

\begin{figure*}[t]
    \centering
    \begin{tcolorbox}
    {\fontfamily{qpl}\selectfont
\begin{enumerate}
\setlength\itemsep{0.1em}
    \item \textunderscore founded \{\textit{source name}\}
\item \textunderscore hosts \{\textit{source name}\}
\item \textunderscore is the founder of \{\textit{source name}\}
\item \textunderscore is the ceo of \{\textit{source name}\}
\item \textunderscore leads \{\textit{source name}\}
\item \textunderscore owns \{\textit{source name}\}
\item \textunderscore publishes \{\textit{source name}\}
\item \textunderscore is the owner of`\{\textit{source name}\}
\item \textunderscore bought \{\textit{source name}\}
\item \textunderscore acquired \{\textit{source name}\}
\item \{\textit{source name}\} is funded through \textunderscore
\item The remit of \{\textit{source name}\} is to \textunderscore
\item The usual audience of \{\textit{source name}\} is \textunderscore
\item \{\textit{source name}\} was awarded \textunderscore
\item \{\textit{source name}\} tends to the \textunderscore
\item Other sources have commented on \{\textit{source name}\}, stating that it tends to be \textunderscore
\item The factuality of \{\textit{source name}\} is reported to be \textunderscore
\item \{\textit{source name}\} failed a fact-check for an article titled ``\textunderscore''
\item \{\textit{source name}\} printed a retraction after failing a fact-check for an article titled ``\textunderscore''
\item \{\textit{source name}\} chose not to cover \textunderscore
\item \{\textit{source name}\} provides original content written by staff journalists
\item \{\textit{source name}\} relies on advertising for revenue
\item \{\textit{source name}\} relies on subscriptions for revenue
\item \{\textit{source name}\} relies on donations for revenue
\item \{\textit{source name}\} has received donations from \textunderscore
\item When aggregating stories, \{\textit{source name}\} relies on information from \textunderscore
\item \{\textit{source name}\} has endorsed \textunderscore
\item \{\textit{source name}\} has an editorial bias towards \textunderscore
\item \textunderscore is the editor of \{\textit{source name}\}
\item \{\textit{source name}\} was fined \textunderscore\ for \textunderscore
\item \{\textit{source name}\} printed a biased article titled ``\textunderscore''
\item \{\textit{source name}\} printed a factually misleading article titled ``\textunderscore''
\item \{\textit{source name}\} had to apologize for \textunderscore
\item \{\textit{source name}\} paid damages to \textunderscore after \textunderscore
\item \{\textit{source name}\} pretends to be \textunderscore
\item It is unknown who \textunderscore
\item \{\textit{source name}\}'s headquarter is located in \textunderscore
\item \{\textit{source name}\} is funded by the \textunderscore\ government
\item \{\textit{source name}\} is a \textunderscore
\item \{\textit{source name}\} uses a peer review process
\item \{\textit{source name}\} uses an internal fact-checking process
\item \{\textit{source name}\} covers the following topics: \textunderscore
\end{enumerate}
}
\end{tcolorbox}
    \caption{Templates used for generating atomic facts in our evaluation setup. Each template is filled with information from the gold background check using \textit{gpt-3.5-turbo} (by replacing the underscore). Templates which are not entailed (or contradicted) by the gold background check are discarded.}
    \label{figure:atomic_fact_templates}
\end{figure*}

\section{Facts in Media Background Checks}
\label{appendix:facts_in_mbcs}

To understand which facts are represented in the MB/FC dataset, we manually analyse 20 randomly selected background checks. We include our findings in Table~\ref{table:facts_appearing}. We note that this is not a complete list -- picking specific high-quality background checks, e.g. background checks for the Guardian, Fox News, or Breitbart, reveals usage of rarer facts. As such, the list of atomic facts we use for evaluation is longer (see Appendix~\ref{appendix:atomic_fact_templates}).

\begin{table*}[t]
\centering
\begin{tabular}{lr}
\toprule
Fact               & Percentage of MBCs \\ \midrule
General bias of \{\textit{source name}\}       & 80\%               \\
Funding model for \{\textit{source name}\}            & 75\%               \\
Remit of \{\textit{source name}\}             & 75\%               \\
Fact-checking history of \{\textit{source name}\} & 65\%               \\
Owner of \{\textit{source name}\}             & 60\%               \\
Publisher of \{\textit{source name}\} & 50\%               \\
Examples of biased articles from \{\textit{source name}\}            & 50\%               \\
Geographical focus of \{\textit{source name}\}             & 50\%               \\
Other history of \{\textit{source name}\}            & 45\%               \\
Multihop Facts          & 40\%               \\
\{\textit{source name}\} does not disclose important information & 35\%               \\
Sources used by \{\textit{source name}\} (e.g., AP)             & 35\%               \\
Founder of \{\textit{source name}\}            & 25\%               \\
Loaded language used by \{\textit{source name}\}             & 25\%               \\
Examples of articles from \{\textit{source name}\} for demonstrating aspects other than bias            & 20\%               \\
Political endorsements by \{\textit{source name}\}              & 10\%               \\
\{\textit{source name}\} masquerades as & 10\%               \\
Awards given to \{\textit{source name}\}            & 10\%               \\
Editor of \{\textit{source name}\}             & 10\%               \\
Bias rating by other site (e.g., NewsGuard)            & 10\%               \\
Demonstration of agreement by \{\textit{source name}\} with scientific consensus         & 5\%               \\
Comparison to other media            & 5\%               \\
\bottomrule
\end{tabular}
\caption{Facts appearing in 20 randomly sampled background checks from Media Bias / Fact Check.}
\label{table:facts_appearing}
\end{table*}

\section{Information-seeking Questions}
\label{appendix:information_seeking_questions}

When retrieving information, we first retrieved documents via the Google Search API. Then, we use a trained question-answering model to select the most salient substrings. The queries and questions used for both can be found in Figure~\ref{figure:information-seeking questions}. 

\begin{figure*}[t]
    \centering
    \begin{tcolorbox}
    {\fontfamily{qpl}\selectfont
\begin{enumerate}
    \item ``\{\textit{source name}\}'' ownership / Who owns ``\{\textit{source name}\}''?
    \item ``\{\textit{source name}\}'' funding / How is ``\{\textit{source name}\}'' funded?
    \item ``\{\textit{source name}\}'' about / What is ``\{\textit{source name}\}''?
    \item ``\{\textit{source name}\}'' political leaning / What is the political leaning of ``\{\textit{source name}\}''?
    \item ``\{\textit{source name}\}'' fact-check / Has ``\{\textit{source name}\}'' failed any fact-checks?
    \item ``\{\textit{source name}\}'' retracted article / Has ``\{\textit{source name}\}'' retracted any articles?
\end{enumerate}
}
\end{tcolorbox}
    \caption{Information-seeking queries input to the Google search API in order to find relevant information on the source. Each query also has a corresponding question, which we use to retrieved the most relevant substring of the search results based on DeBERTa~\citep{he2021deberta}.}
    \label{figure:information-seeking questions}
\end{figure*}

\section{Questionnaire}

\subsection{Part 1}
\label{appendix:survey_part_1}

\begin{figure*}
\centering
\includegraphics[scale=0.5]{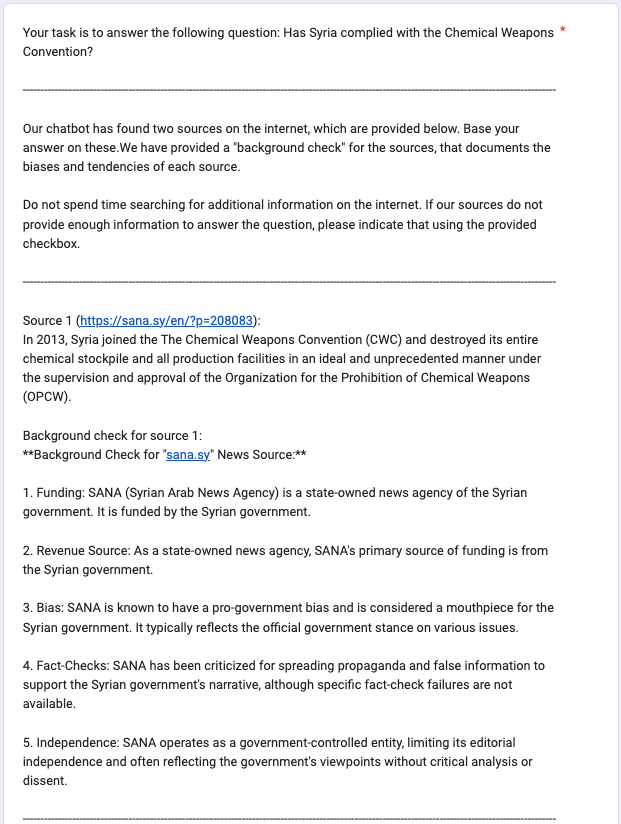}
\caption{An example page from the first half of our human evaluation questionnaire. Given a controversial question, two sources, and their background checks, our annotators were asked to answer the question. They were furthermore asked to answer if they had sufficient information, and how difficult the task were. Continued in Figures~\ref{fig:survey2} and \ref{fig:survey3}.}
\label{fig:survey1}
\end{figure*}

\begin{figure*}
\centering
\includegraphics[scale=0.5]{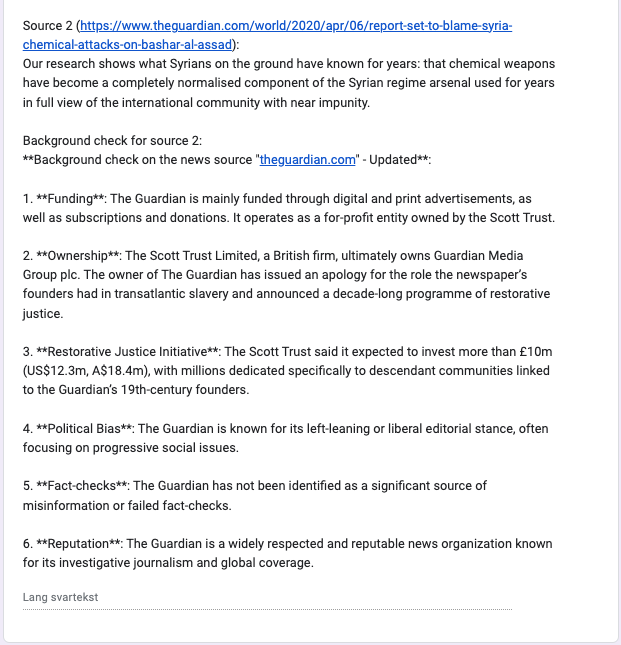}
\caption{An example page from the first half of our human evaluation questionnaire. Given a controversial question, two sources, and their background checks, our annotators were asked to answer the question. They were furthermore asked to answer if they had sufficient information, and how difficult the task were. Continued from Figure~\ref{fig:survey1}.}
\label{fig:survey2}
\end{figure*}

\begin{figure*}
\centering
\includegraphics[scale=0.5]{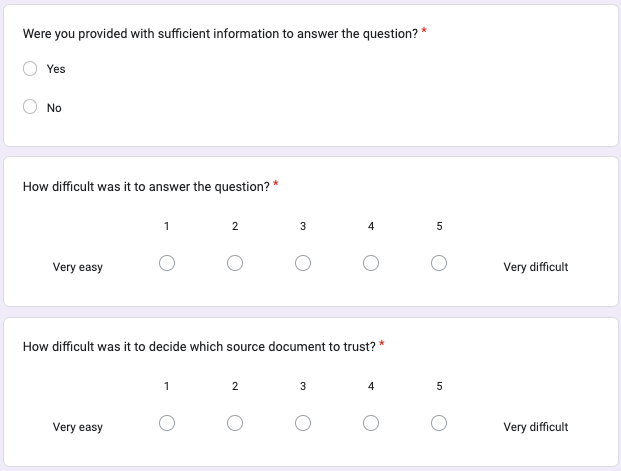}
\caption{An example page from the first half of our human evaluation questionnaire. Given a controversial question, two sources, and their background checks, our annotators were asked to answer the question. They were furthermore asked to answer if they had sufficient information, and how difficult the task were. Continued from Figure~\ref{fig:survey2}.}
\label{fig:survey3}
\end{figure*}

The questionnaire pages used in the first half of our survey can be seen in Figures~\ref{fig:survey1},~\ref{fig:survey2}, and~\ref{fig:survey3}. These focus on evaluating if MBCs are helpful for \textit{humans} when creating meaning from untrustworthy evidence documents.

\subsection{Part 2}
\label{appendix:survey_part_2}

\begin{figure*}
\centering
\includegraphics[scale=0.5]{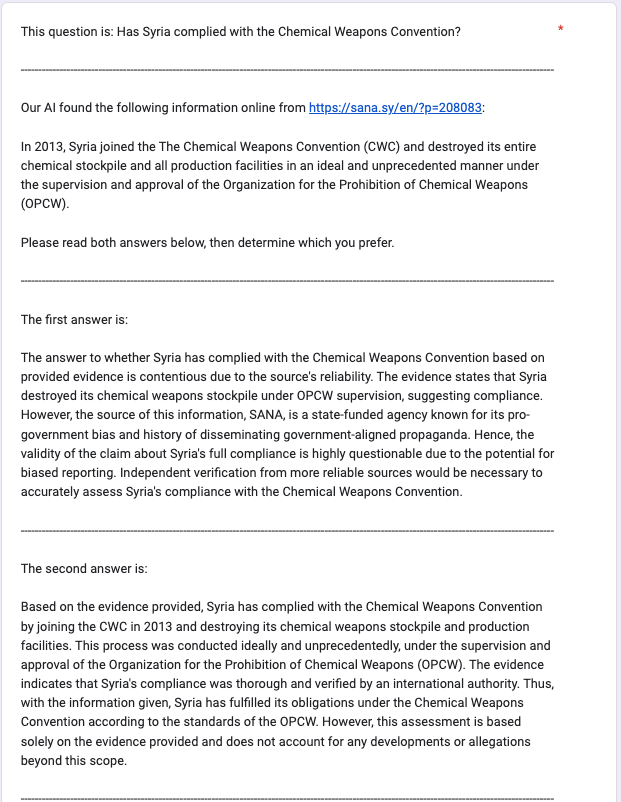}
\caption{An example page from the second half of our human evaluation questionnaire. Given answers provided by GPT-4 with and without an MBC, annotators were asked for their preferences, as well as whether any answer was misleading. Continued in Figures~\ref{fig:survey5}.}
\label{fig:survey4}
\end{figure*}

\begin{figure*}
\centering
\includegraphics[scale=0.5]{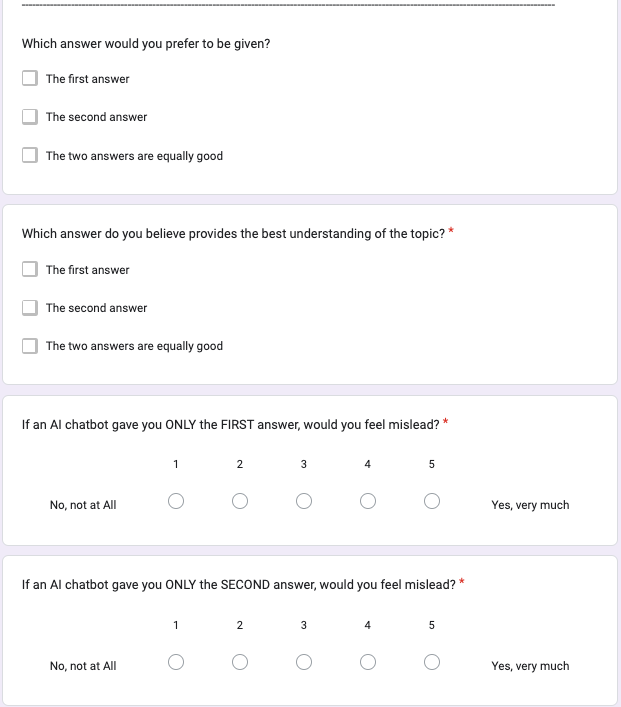}
\caption{An example page from the second half of our human evaluation questionnaire. Given answers provided by GPT-4 with and without an MBC, annotators were asked for their preferences, as well as whether any answer was misleading. Continued from Figure~\ref{fig:survey4}.}
\label{fig:survey5}
\end{figure*}

The questionnaire pages used in the second half of our survey can be seen in Figures~\ref{fig:survey4} and~\ref{fig:survey5}. These focus on evaluating if MBCs are helpful for \textit{NLP models} when creating meaning from untrustworthy evidence documents.

\section{Data Statement}

Following \citet{bender-friedman-2018-data}, we include a data statement describing the characteristics of MB/FC.

\subsection{Curation Rationale}
\label{appendix:curation}
We processed a total of 6,709 source documents from the Media Bias / Fact Check website, which provides volunteer-written assessments of English-language media sources. The intended use is to provide empirical evidence for the level of source-critical analysis possible for NLP models, as well as to drive research into new tools that enable source-critical analysis by models and humans.

\subsection{Language variety}
As MB/FC only provides analysis of English-language sources, in English, the same holds true for our dataset.

\subsection{Speaker demographics}

We did not analyse the demographics of the individual writers and editors for each media source.

\subsection{Annotator demographics}
For this dataset, we did not rely on human annotators beyond our own group. We processed data from Media Bias / Fact Check, a volunteer-driven project, with details available here: \url{https://mediabiasfactcheck.com/frequently-asked-questions/}.

\subsection{Speech situation}
The MBCs included in this dataset were provided by the Media Bias / Fact Check volunteers for the purpose of educating the public on media bias and deceptive news practises.

\subsection{Text characteristics}
We compute various statistics for the text included in this dataset; see Section~\ref{section:dataset}.

\end{document}